\documentclass[conference]{IEEEtran}
\IEEEoverridecommandlockouts
\usepackage{cite}
\usepackage{amsmath,amssymb,amsfonts}
\usepackage{algorithmic}
\usepackage{graphicx}
\usepackage{textcomp}
\usepackage{xcolor}
\usepackage{tabularx}
\usepackage{colortbl} 
\usepackage{booktabs}
\usepackage{multirow}
\usepackage[table]{xcolor}
\usepackage[normalem]{ulem}
\usepackage{adjustbox}
\usepackage{algorithm}
\usepackage{cite}

\definecolor{cvprblue}{rgb}{0.21,0.49,0.74}
\definecolor{lightblue}{RGB}{200, 220, 255}
\definecolor{midblue}{RGB}{140, 180, 250}
\definecolor{darkblue}{RGB}{90, 130, 230}
\definecolor{darkred}{RGB}{220,20,60}
\usepackage[pagebackref,breaklinks,colorlinks,allcolors=cvprblue]{hyperref}

\usepackage[title]{appendix}

\begin{document}

\title{Latent Anomaly Knowledge Excavation: Unveiling Sparse Sensitive Neurons in Vision-Language Models
}

\author{
Shaotian Li$^{1}$ \quad
Shangze Li$^{2}$ \quad
Chuancheng Shi$^{3}$ \quad
Wenhua Wu$^{3}$ \quad
Yanqiu Wu$^{1}$ \quad \\
Xiaohan Yu$^{1}$ \quad
Fei Shen$^{4\dagger}$ \quad
Tat-Seng Chua$^{4}$\\[4pt]
$^{1}$ Macquarie University \quad
$^{2}$ Nanjing University of Science and Technology \quad \\
$^{3}$ The University of Sydney \quad
$^{4}$ National University of Singapore\\
$^{\dagger}$ Corresponding Author
}

\maketitle

\begin{abstract}
Large-scale vision-language models (VLMs) exhibit remarkable zero-shot capabilities, yet the internal mechanisms driving their anomaly detection (AD) performance remain poorly understood. Current methods predominantly treat VLMs as black-box feature extractors, assuming that anomaly-specific knowledge must be acquired through external adapters or memory banks. In this paper, we challenge this assumption by arguing that anomaly knowledge is intrinsically embedded within pre-trained models but remains latent and under-activated. We hypothesize that this knowledge is concentrated within a sparse subset of anomaly-sensitive neurons. To validate this, we propose latent anomaly knowledge excavation (LAKE), a training-free framework that identifies and elicits these critical neuronal signals using only a minimal set of normal samples. By isolating these sensitive neurons, LAKE constructs a highly compact normality representation that integrates visual structural deviations with cross-modal semantic activations. Extensive experiments on industrial AD benchmarks demonstrate that LAKE achieves state-of-the-art performance while providing intrinsic, neuron-level interpretability. Ultimately, our work advocates for a paradigm shift: redefining anomaly detection as the targeted activation of latent pre-trained knowledge rather than the acquisition of a downstream task.
\end{abstract}


\section{Introduction}

Large-scale pre-trained vision-language models (VLMs) have demonstrated remarkable capabilities in cross-modal alignment, open-vocabulary recognition, and zero-shot generalization \cite{radford2021learningtransferablevisualmodels, caron2021emerging, jia2021scalingvisualvisionlanguagerepresentation, liu2023llava}. This success suggests that beyond merely learning visual appearances and semantic concepts, these models have internalized deep structural knowledge of the boundaries of normality. Consequently, a pivotal scientific question emerges: Does the logical reasoning required for anomaly detection necessitate explicit activation through downstream fine-tuning, or has it already undergone spontaneous semantic precipitation as high-dimensional feature distributions during massive self-supervised pre-training? Defining these internal knowledge boundaries is essential for engineering truly universal and reliable anomaly detection systems.

Despite this potential, existing zero- and few-shot anomaly detection methods \cite{cai2025medianomaly, zhang2025eiad} predominantly treat these powerful VLMs as opaque feature extractors. Through external adaptations such as prompt tuning \cite{zhouanomalyclip, cao2024adaclip, cai2025towards}, auxiliary adapters \cite{gao2026adaptclip}, memory banks \cite{roth2022totalrecallindustrialanomaly, xu2026mradzeroshotanomalydetection, Ni_2025_BMVC}, or synthetic anomaly generation \cite{zhang2024realnet, jiang2026anomagic, jin2025dual}, these approaches rely on the premise that anomaly perception must be explicitly shaped by downstream modules rather than originating from internal knowledge. While these external modeling techniques improve empirical performance, they inherently bypass the internal dynamics of the model, leaving the innate discriminative potential of the VLM's latent space entirely unexamined.
This black-box paradigm introduces two fundamental limitations. First, current methods primarily perform compression, alignment, and matching at the macro-feature level, failing to reveal the mechanistic origins of zero- and few-shot anomaly detection capabilities \cite{golovanevsky2025vlmsnoticemechanisticinterpretability}. Second, existing explainable methods in this domain are largely restricted to post-hoc attribution \cite{benou2025tellvisuallyexplainabledeep} or language-based diagnostics \cite{zhang2025eiad, jin2025logicad}. They may explain why a model made a specific prediction, but they cannot identify which internal units actually carry the anomaly knowledge \cite{huo-etal-2024-mmneuron}. Therefore, the field still lacks an anomaly understanding framework grounded in internal model mechanics.

\begin{figure}[t]
\centering
\includegraphics[width=1\linewidth]{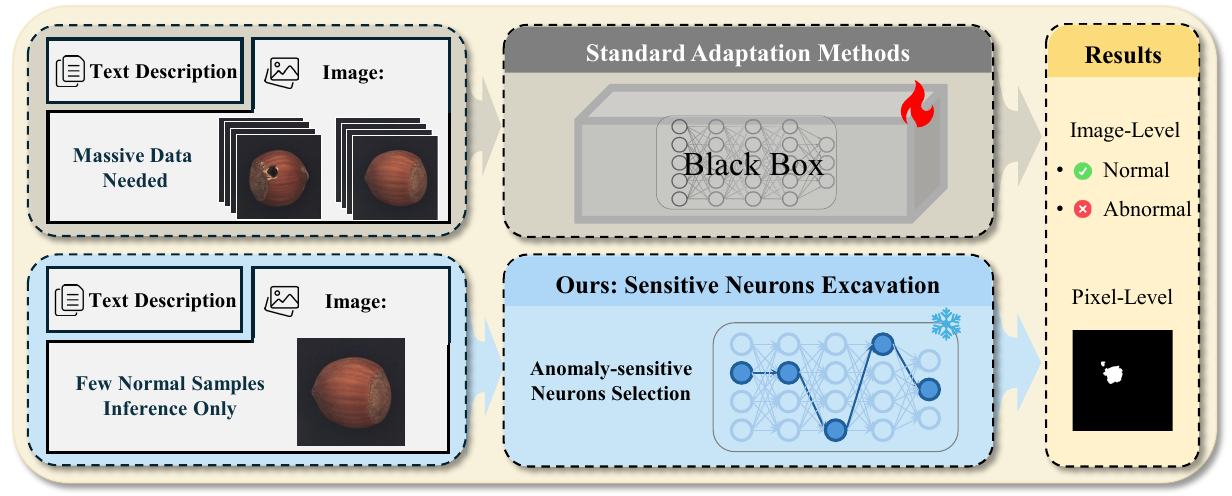} 
\vspace{-0.7cm}
\caption{{Comparison of anomaly detection paradigms.} Unlike existing approaches that rely on external black-box modeling, our method intrinsically elicits anomaly awareness by isolating and activating sparse sensitive neurons within the frozen network.}
\label{fig:ques}
\vspace{-0.5cm}
\end{figure}

To bridge this gap, we hypothesize that anomaly knowledge is not absent from pre-trained VLMs; rather, it remains latent and under-activated during standard inference. Drawing inspiration from recent findings on knowledge localization in foundation models \cite{dai2022knowledge, meng2022locating, gandelsman2024interpreting}, we propose that the model's intrinsic sensitivity to subtle visual deviations and semantic divergence from expected normality is not uniformly distributed across the entire network \cite{shi2025culture}. Instead, it is concentrated within a sparse set of functionally sensitive neurons. By utilizing a small number of normal support samples to identify and activate these critical neurons, it is possible to directly elicit the model's latent anomaly knowledge without modifying its underlying architecture, as shown in Figure \ref{fig:ques}.

To validate this hypothesis, we introduce latent anomaly knowledge excavation (LAKE), an interpretable framework that directly elicits this capability. By profiling the distributional variance of normal samples, our framework localizes a subset of anomaly-sensitive neurons to construct a highly compact visual memory representation, drastically reducing the massive feature redundancy seen in standard external memory banks \cite{dou2026dna, he2025rareclip, xu2026mradzeroshotanomalydetection}. Simultaneously, LAKE leverages cross-modal textual activation to explicitly probe anomalous regions at the semantic level. Diverging from traditional adapter-dependent tuning routes \cite{gao2026adaptclip, ma2025aaclipenhancingzeroshotanomaly, shiri2025madclip}, our method fundamentally reformulates anomaly detection as the elegant process of excavating, localizing, and activating latent anomaly knowledge natively within the pre-trained model.
The significance of this perspective extends beyond empirical performance gains. It establishes a novel, mechanistically interpretable framework for anomaly understanding. In summary, our main contributions are as follows:

\begin{itemize}
    \item We reframe few-shot anomaly detection by demonstrating that anomaly-discriminative capabilities can be directly elicited from the sparse internal neurons of frozen VLMs, bypassing the need for heavy external adapters or massive memory banks.
    \item We introduce LAKE, which seamlessly combines variance-based neuron localization with cross-modal textual activation to achieve intrinsic, microscopic-level interpretability without relying on post-hoc attribution.
    \item Extensive experiments on industrial benchmarks prove that LAKE simultaneously achieves superior detection accuracy and fine-grained interpretability. Furthermore, evaluations on medical datasets confirm strong cross-domain generalization for high-reliability visual applications.
\end{itemize}

\section{Related Work}
\noindent\textbf{Anomaly Detection.} 
Anomaly detection (AD) aims to identify localized deviations without anomaly supervision. Traditional memory-based methods (e.g., PatchCore~\cite{roth2022totalrecallindustrialanomaly}, MRAD~\cite{xu2026mradzeroshotanomalydetection}) and their efficient coreset variants (e.g., CFA, FSLC~\cite{Ni_2025_BMVC}) rely on external reference galleries for feature matching. However, they operate at the macro-feature level, failing to reveal which internal dimensions inherently discriminate anomalies. Recently, Vision-Language Models (VLMs) have dominated zero-shot AD. While early approaches like WinCLIP~\cite{Jeong_2023_CVPR} use hand-crafted prompts, recent state-of-the-art typically rely on learnable prompts or lightweight adapters (e.g., AnomalyCLIP~\cite{zhouanomalyclip}, AdaCLIP~\cite{cao2024adaclip}, AdaptCLIP~\cite{gao2026adaptclip}, AA-CLIP~\cite{ma2025aaclipenhancingzeroshotanomaly}), visual tokens (VisualAD~\cite{hou2026visualadlanguagefreezeroshotanomaly}), or synthetic generation (RealNet~\cite{zhang2024realnet}). In parallel, MLLMs~\cite{li2026iad, xu2025towards, chen2025can} further enhance diagnosis via textual reasoning. Despite their success, these approaches predominantly treat pre-trained models as black boxes. By relying heavily on external adaptation, memory construction, or post-hoc attribution, they bypass the fundamental question of whether anomaly knowledge is already natively encoded within the frozen model.

\noindent\textbf{Latent Knowledge Excavation and Neuron Interpretability.}
Mechanistic interpretability reveals that foundation models are not opaque black boxes; rather, their capabilities can be localized to sparse, functionally specialized units. Because these features are typically entangled within polysemantic activation spaces, some approaches employ sparse autoencoders (SAEs) to decompose representations into interpretable concepts~\cite{lou2025saevinterpretingmultimodalmodels, zaigrajew2025interpretingcliphierarchicalsparse, huang2025tidetemporalawaresparse, lim2025sparseautoencodersrevealselective}. However, SAEs require resource-intensive auxiliary training and are not tailored for anomaly detection (AD). Alternatively, research on concept circuits and activation patching demonstrates that specific semantic behaviors stem from localized subnetworks~\cite{kwon2025granularconceptcircuitsfinegrained, huo-etal-2024-mmneuron, cui2026llms, golovanevsky2025vlmsnoticemechanisticinterpretability}. These studies prove that latent knowledge already exists within pre-trained models and can be extracted or steered without conventional downstream fine-tuning~\cite{shi2025culture, shi2026tracerouter, dou2026dna, fang2024towards}. Yet, this training-free excavation perspective remains largely unexplored in industrial AD. To bridge this gap, LAKE mathematically isolates anomaly-sensitive neurons to awaken intrinsically encoded anomaly awareness. This explicitly reformulates AD from external task adaptation to intrinsic latent knowledge excavation.

\begin{figure*}[t]
\centering
\includegraphics[width=0.93\linewidth]{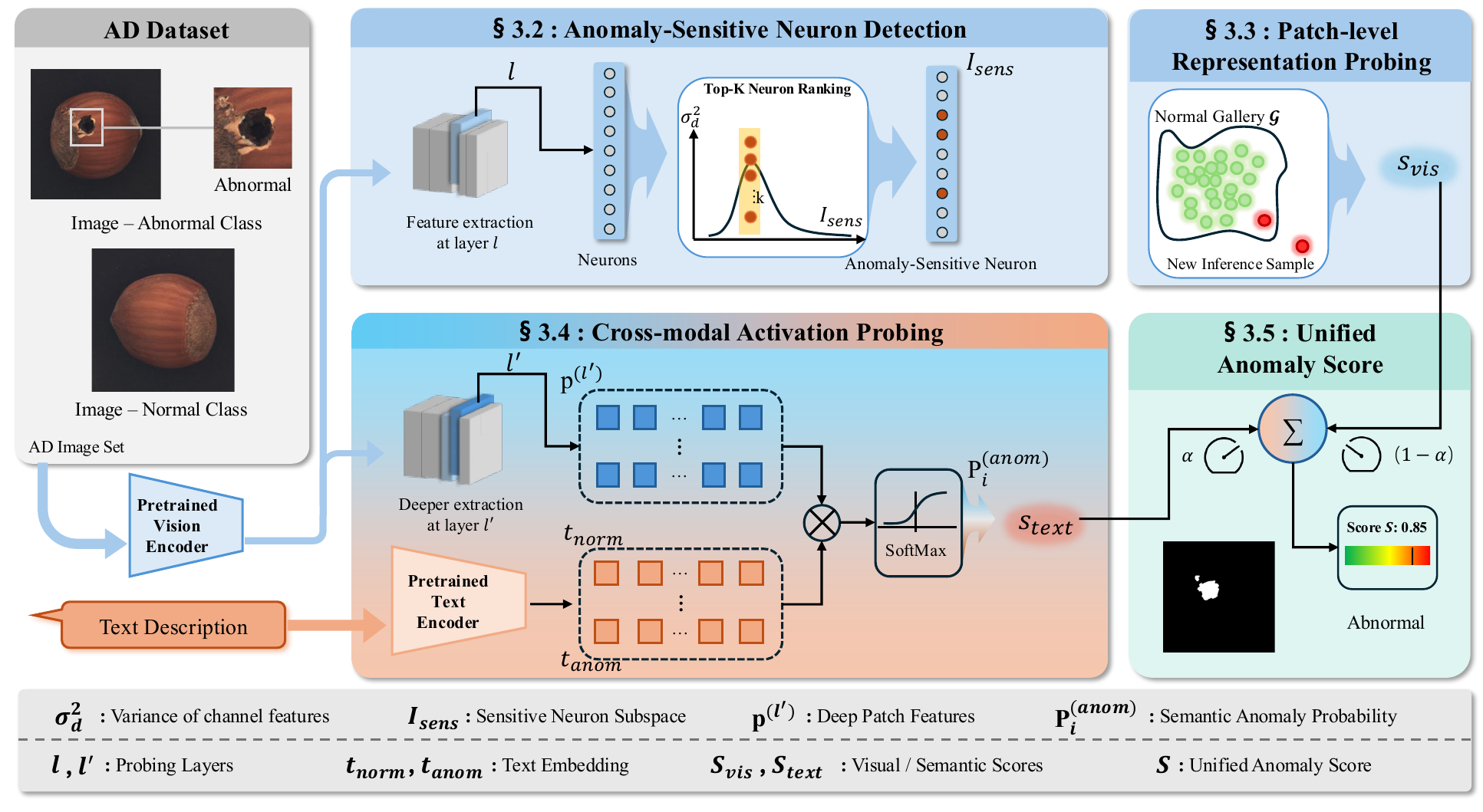}
\vspace{-0.4cm}

\caption{{The LAKE framework.} Features from layer $l$ are projected into a variance-filtered sensitive subspace ($I_{sens}$) and compared against a normal gallery ($\mathcal{G}$) to compute the visual score ($S_{vis}$). Simultaneously, deeper features ($P^{(l')}$) are aligned with text embeddings ($t_{norm}, t_{anom}$) to extract the semantic score ($S_{text}$). Both scores are fused via weight $\alpha$ to produce the unified anomaly score ($S$).}
\label{fig:pipeline}
\vspace{-0.4cm}
\end{figure*}

\section{Method}
We hypothesize that pre-trained VLMs intrinsically encode anomaly knowledge that remains under-activated during standard inference. Instead of relying on external adaptations, our proposed LAKE framework directly probes the frozen model internally to excavate this latent knowledge. As shown in Figure~\ref{fig:pipeline} and Algorithm~\ref{alg:lake}, LAKE consists of three core steps: (1) identifying anomaly-sensitive neurons from a small normal support set, (2) probing patch-level visual deviations within this sensitive subspace, and (3) verifying semantic abnormality via cross-modal activation. Ultimately, LAKE fuses structural and semantic cues to enable robust, training-free, and interpretable anomaly detection.

\subsection{Problem Setup}

Given an input image $x$, we extract patch-level visual features from a frozen vision encoder at layer $l$:
\begin{equation}
\mathbf{H}^{(l)}(x)
=
\left[
\mathbf{h}^{(l)}_1(x),\mathbf{h}^{(l)}_2(x),\dots,\mathbf{h}^{(l)}_N(x)
\right]
\in \mathbb{R}^{N \times D},
\end{equation}
where $N$ is the number of spatial tokens and $D$ is the feature dimension. Here, $\mathbf{h}^{(l)}_i(x) \in \mathbb{R}^{D}$ denotes the feature of the $i$-th token.

We assume access only to a small normal support set
\begin{equation}
\mathcal{X}_{\mathrm{norm}} = \{x^{(1)},x^{(2)},\dots,x^{(M)}\},
\end{equation}
where all images are anomaly-free. No anomalous samples, additional optimization, or parameter tuning are used. The task is to determine whether a test image departs from the normal distribution and, if so, where such abnormality is localized.

The key challenge is that not all hidden dimensions contribute equally to anomaly perception. Many channels encode generic appearance or redundant background statistics, while only a small subset may respond strongly to subtle abnormal structures. Therefore, rather than comparing all feature dimensions, LAKE first identifies a sparse anomaly-sensitive subspace from normal data and then performs visual-semantic probing within this subspace.

\subsection{Anomaly-Sensitive Neuron Detection}
Our first step is to explicitly identify which feature dimensions are most informative for anomaly discrimination. In traditional feature spaces, anomalies are often assumed to reside in low-variance residual directions. However, within the dense, polysemantic activation space of foundation models, neurons with near-zero variance on normal data typically correspond to dormant concepts or task-irrelevant background noise. Conversely, channels exhibiting structured, high variance under normal data approximate the principal axes of the specific target manifold. These active neurons continuously encode the core structural and semantic components of the expected normality. When an anomaly occurs, it fundamentally disrupts this underlying normal manifold. Therefore, the effects of such disruptions are most saliently manifested as deviations along these active, high-variance directions, rather than as sudden spikes in dormant neurons.
To quantify this, we compute the variance of each channel over all normal support samples and all spatial tokens:
\begin{equation}
\sigma_d^2
=
\operatorname{Var}_{x \in \mathcal{X}_{\mathrm{norm}},\, i \in \{1,\dots,N\}}
\left(
h^{(l)}_{i,d}(x)
\right),
\label{eq:variance}
\end{equation}
where $h^{(l)}_{i,d}(x)$ is the $d$-th entry of $\mathbf{h}^{(l)}_i(x)$. We then rank all channels according to $\sigma_d^2$ and select the top-$K$ dimensions:
\begin{equation}
\mathcal{I}_{\mathrm{sens}}
=
\operatorname{TopK}\left(\{\sigma_d^2\}_{d=1}^{D}, K\right),
\label{eq:topk}
\end{equation}
where $\mathcal{I}_{\mathrm{sens}} \subset \{1,\dots,D\}$ and $|\mathcal{I}_{\mathrm{sens}}| = K$.

This set $\mathcal{I}_{\mathrm{sens}}$ defines our \textit{anomaly-sensitive neuron subspace}. It provides a compact internal representation of the most responsive feature dimensions under normal data, effectively bypassing the massive feature redundancy typical of foundation models, and forms the basis of the subsequent probing stages. As shown in Algorithm~\ref{alg:lake}, this subspace is estimated only once from the normal support set and reused during inference, keeping the overall framework fully training-free.

\subsection{Patch-level Representation Probing}

After identifying $\mathcal{I}_{\mathrm{sens}}$, we next ask how anomalies manifest spatially. Since many industrial defects are local and occupy only small regions, it is not sufficient to reason at the image level alone. We instead project each patch token into the sensitive subspace and explicitly measure how far it deviates from normal patterns.

For each patch token, we define its projected representation as
\begin{equation}
\mathbf{z}_i(x)
=
\mathbf{h}^{(l)}_i(x)[\mathcal{I}_{\mathrm{sens}}]
\in \mathbb{R}^{K}.
\label{eq:projection}
\end{equation}
Using all normal support images, we construct a reference gallery:
\begin{equation}
\mathcal{G}
=
\left\{
\mathbf{z}_j(x)\;|\;x \in \mathcal{X}_{\mathrm{norm}},\; j \in \{1,\dots,N\}
\right\}.
\label{eq:gallery}
\end{equation}
This gallery can be viewed as a discrete approximation of the normal manifold in the sensitive subspace.

For a test image $x$, we measure the anomaly deviation of token $i$ by its nearest-neighbor distance to the gallery:
\begin{equation}
d_i(x)
=
\min_{\mathbf{g} \in \mathcal{G}}
\frac{1}{2}
\left(
1-
\frac{\mathbf{z}_i(x)^\top \mathbf{g}}
{\|\mathbf{z}_i(x)\|_2 \, \|\mathbf{g}\|_2}
\right).
\label{eq:nn_distance}
\end{equation}
This distance is small when the test token lies close to the normal manifold and increases when it deviates from normal appearance patterns. We then aggregate all token deviations by max pooling:
\begin{equation}
S_{\mathrm{vis}}(x)
=
\max_{i \in \{1,\dots,N\}} d_i(x).
\label{eq:svis}
\end{equation}

The use of max pooling is important. In anomaly detection, abnormal regions are often sparse and can be easily suppressed by global averaging. By keeping the maximum deviation, LAKE preserves the most salient abnormal evidence and naturally supports fine-grained localization. In Algorithm~\ref{alg:lake}, this stage corresponds to gallery construction from the support set and nearest-neighbor deviation computation on the test image.

\subsection{Cross-modal Activation Probing}

Although the visual probing stage captures geometric departure from normality, it does not explicitly answer whether such deviation is semantically abnormal. For example, strong visual change may arise from benign texture variation rather than an actual defect. To provide semantic verification, we introduce a second probing stage that activates anomaly knowledge through cross-modal alignment.
Specifically, we extract deeper patch representations from layer $l'$:
\begin{equation}
\mathbf{P}^{(l')}(x)
=
\left[
\mathbf{p}^{(l')}_1(x),\mathbf{p}^{(l')}_2(x),\dots,\mathbf{p}^{(l')}_N(x)
\right]
\in \mathbb{R}^{N \times D_t},
\end{equation}
where $\mathbf{p}^{(l')}_i(x) \in \mathbb{R}^{D_t}$ lies in the same embedding space as the text encoder. We then encode two textual prompts describing the normal and anomalous states:
\begin{equation}
\mathbf{t}_{\mathrm{norm}},\mathbf{t}_{\mathrm{anom}} \in \mathbb{R}^{D_t}.
\end{equation}
For each token, we compute its similarity to both prompts:
\begin{equation}
\mathbf{s}_i(x)
=
\begin{bmatrix}
\mathbf{p}^{(l')}_i(x)^\top \mathbf{t}_{\mathrm{norm}} \\
\mathbf{p}^{(l')}_i(x)^\top \mathbf{t}_{\mathrm{anom}}
\end{bmatrix}
\in \mathbb{R}^{2},
\label{eq:text_score}
\end{equation}
followed by
\begin{equation}
\mathbf{P}_i(x)
=
\operatorname{softmax}\big(\mathbf{s}_i(x)\big).
\label{eq:softmax}
\end{equation}
The semantic anomaly score is defined as
\begin{equation}
S_{\mathrm{text}}(x)
=
\max_{i \in \{1,\dots,N\}}
P_i^{(\mathrm{anom})}(x).
\label{eq:stext}
\end{equation}

This cross-modal activation stage complements Eq.~\eqref{eq:svis} in an important way. The visual branch asks whether a region departs from the normal manifold, whereas the semantic branch asks whether that region aligns with the concept of abnormality. Their combination allows LAKE to reduce false positives caused by purely geometric variation while maintaining sensitivity to subtle defects. In Algorithm~\ref{alg:lake}, this stage computes text similarities and token-wise anomalous probabilities from deeper visual features.

\begin{algorithm}[t]
\caption{Inference pipeline of LAKE}
\label{alg:lake}
\begin{algorithmic}[1]
\REQUIRE Frozen VLM, normal support set $\mathcal{X}_{\mathrm{norm}}$, test image $x$, probing layers $l,l'$, subspace size $K$, fusion weight $\alpha$
\ENSURE Final anomaly score $S(x)$

\STATE Extract support features $\mathbf{H}^{(l)}(x')$ for each $x' \in \mathcal{X}_{\mathrm{norm}}$
\FOR{$d=1$ to $D$}
    \STATE Compute channel variance $\sigma_d^2$ using Eq.~\eqref{eq:variance}
\ENDFOR
\STATE Select anomaly-sensitive subspace $\mathcal{I}_{\mathrm{sens}}$ using Eq.~\eqref{eq:topk}

\STATE Initialize gallery $\mathcal{G} \leftarrow \emptyset$
\FOR{each $x' \in \mathcal{X}_{\mathrm{norm}}$}
    \FOR{each token $j \in \{1,\dots,N\}$}
        \STATE Compute projected feature $\mathbf{z}_j(x')$ using Eq.~\eqref{eq:projection}
        \STATE Add $\mathbf{z}_j(x')$ to $\mathcal{G}$
    \ENDFOR
\ENDFOR

\STATE Extract test features $\mathbf{H}^{(l)}(x)$ and compute $\mathbf{z}_i(x)$ using Eq.~\eqref{eq:projection}
\FOR{each token $i \in \{1,\dots,N\}$}
    \STATE Compute visual deviation $d_i(x)$ using Eq.~\eqref{eq:nn_distance}
\ENDFOR
\STATE Compute $S_{\mathrm{vis}}(x)$ using Eq.~\eqref{eq:svis}

\STATE Extract deeper test features $\mathbf{P}^{(l')}(x)$
\STATE Encode $\mathbf{t}_{\mathrm{norm}}$ and $\mathbf{t}_{\mathrm{anom}}$
\FOR{each token $i \in \{1,\dots,N\}$}
    \STATE Compute similarity $\mathbf{s}_i(x)$ using Eq.~\eqref{eq:text_score}
    \STATE Compute probability $\mathbf{P}_i(x)$ using Eq.~\eqref{eq:softmax}
\ENDFOR
\STATE Compute $S_{\mathrm{text}}(x)$ using Eq.~\eqref{eq:stext}

\STATE Compute final score $S(x)$ using Eq.~\eqref{eq:final_score}
\RETURN $S(x)$
\end{algorithmic}
\end{algorithm}

\begin{table*}[t]
  \centering
\caption{Image-level anomaly detection results on the MVTec-AD, VisA, and BTAD benchmarks. Evaluation metrics include AUROC, $F_1$-max, and AP. The highest scores are in bold, alongside relative performance gains over the OpenCLIP model.}
  \label{tab:image_level_transposed}
  \renewcommand{\arraystretch}{1} 
  \resizebox{\linewidth}{!}{
    \begin{tabular}{l ccc ccc ccc}
    \toprule
    \multirow{2}{*}{\textbf{Method}} & \multicolumn{3}{c}{\textbf{MVTec-AD~\cite{bergmann2019mvtec}}} & \multicolumn{3}{c}{\textbf{VisA~\cite{zou2022spot}}} & \multicolumn{3}{c}{\textbf{BTAD~\cite{btad2020}}} \\
    
    \cmidrule(lr){2-4} \cmidrule(lr){5-7} \cmidrule(lr){8-10}
    
    & AUROC & $F_1$-max & AP & AUROC & $F_1$-max & AP & AUROC & $F_1$-max & AP \\
    \midrule
    OpenCLIP~\cite{radford2021learning} & 74.1 & 88.5 & 89.1 & 60.2 & 73.0 & 66.2 & 25.7 & 66.0 & 49.8  \\
    \midrule
    
    WinCLIP~\cite{jeong2023winclip}          & 90.4 & 92.7 & 95.6 & 75.6 & 78.2 & 78.8 & 68.2 & 67.8 & 70.9 \\
    CLIP-AD~\cite{chen2024clip}          & 74.1 & 86.3 & 88.1 & 66.2 & 74.3 & 71.4 & 66.7 & 65.9 & 67.3 \\
    AnomalyCLIP~\cite{zhouanomalyclip}      & 91.6 & 92.7 & 96.2 & 81.0 & 80.3 & 84.4 & 88.7 & 86.0 & 90.6 \\
    AdaCLIP~\cite{cao2024adaclip}          & 92.2 & 92.7 & 96.4 & 79.7 & 79.6 & 83.2 & 90.0 & 87.2 & 91.5 \\
    VisualAD (CLIP)~\cite{hou2026visualadlanguagefreezeroshotanomaly}  & 92.2 & 93.2 & 96.7 & 84.7 & 82.5 & 87.6 & 94.9 & 93.9 & 97.0 \\
    VisualAD (DINOv2)~\cite{hou2026visualadlanguagefreezeroshotanomaly}& 90.1 & 92.4 & 94.8 & 83.1 & 81.4 & 86.8 & 88.2 & 84.7 & 89.7 \\
    
    \midrule 
    \rowcolor[HTML]{E5F0F7} 
    \textbf{LAKE (Ours)}    & \textbf{94.7} \textcolor[HTML]{008000}{\scriptsize{(+20.6\%)}} & \textbf{93.9} \textcolor[HTML]{008000}{\scriptsize{(+5.4\%)}} & \textbf{96.8} \textcolor[HTML]{008000}{\scriptsize{(+7.7\%)}} & \textbf{89.4} \textcolor[HTML]{008000}{\scriptsize{(+29.2\%)}} & \textbf{86.2} \textcolor[HTML]{008000}{\scriptsize{(+13.2\%)}} & \textbf{90.0} \textcolor[HTML]{008000}{\scriptsize{(+23.8\%)}} &\textbf{96.2} \textcolor[HTML]{008000}{\scriptsize{(+70.5\%)}} & \textbf{95.0} \textcolor[HTML]{008000}{\scriptsize{(+29.0\%)}} & \textbf{97.2} \textcolor[HTML]{008000}{\scriptsize{(+47.4\%)}} \\
    \bottomrule
    \end{tabular}
  }
\end{table*}

\begin{table*}[t]
  \centering
\caption{Pixel-level anomaly localization results on three datasets. Evaluated metrics are AUROC, $F_1$-max, AP, and PRO. The highest scores are shown in italics, alongside the relative improvements achieved over the OpenCLIP model.}
  \label{tab:pixel_level_transposed}
  \renewcommand{\arraystretch}{1}
  \resizebox{\linewidth}{!}{
   
    \begin{tabular}{l cccc cccc cccc}
    \toprule
    \multirow{2}{*}{\textbf{Method}} & \multicolumn{4}{c}{\textbf{MVTec-AD}~\cite{bergmann2019mvtec}} & \multicolumn{4}{c}{\textbf{VisA~\cite{zou2022spot}}} & \multicolumn{4}{c}{\textbf{BTAD~\cite{btad2020}}} \\
    
    \cmidrule(lr){2-5} \cmidrule(lr){6-9} \cmidrule(lr){10-13}
    
    & AUROC & $F_1$-max & AP & PRO & AUROC & $F_1$-max & AP & PRO & AUROC & $F_1$-max & AP & PRO \\
    \midrule
    OpenCLIP~\cite{radford2021learning}   & 35.6 & 2.8 & -- & 6.9 & 43.6 & 1.5 & -- & 14.0 & 41.2 & 6.8 & -- & 10.8 \\
    \midrule
    
    WinCLIP~\cite{jeong2023winclip}           &  82.3  & 24.8  & 18.2  &  62.0 &  73.2 & 9.0  &  5.4   & 51.1  &  72.7 &  18.5 & 12.9  &  27.3 \\
    CLIP-AD~\cite{chen2024clip}           & 77.9  & 26.3  & 21.1  & 55.7  & 93.0  & 24.1  & 17.9  &  80.2 & 80.9  &  24.1 & 18.3  &  41.4 \\
    AnomalyCLIP~\cite{zhouanomalyclip}      & 91.0  & 38.9  & 34.4  &  81.7 & 95.4  & 27.6  &  20.7 & 86.4 & 93.0  & 47.1  & 41.5  &  71.0 \\
    AdaCLIP~\cite{cao2024adaclip}          & 88.5  & 43.9  & 41.0  &  47.6 &  95.1 & 33.8 &  29.2 & 71.3  & 87.7  & 42.3  & 36.6  &  17.1 \\
    VisualAD (CLIP)~\cite{hou2026visualadlanguagefreezeroshotanomaly}  & 90.8   & 43.9  & 41.2   &  87.5 & 95.8   &  34.6 & 28.4  & 91.0  & 91.1  & 49.8  &  43.1 & 80.4  \\
    VisualAD (DINOv2)~\cite{hou2026visualadlanguagefreezeroshotanomaly} &  91.3 &  47.4 &  45.4 &  88.6 &  95.3 & 35.2  &  29.9 &  88.2 &  93.4 &  42.6 & 38.7  & 76.7  \\
    
    \midrule 
    \rowcolor[HTML]{E5F0F7}
    \textbf{LAKE (Ours)}    & \textbf{93.7} \textcolor[HTML]{008000}{\scriptsize{(+58.1\%)}} & \textbf{50.8} \textcolor[HTML]{008000}{\scriptsize{(+48.0\%)}} & \textbf{45.5} & \textbf{88.9} \textcolor[HTML]{008000}{\scriptsize{(+82.0\%)}} & \textbf{95.7} \textcolor[HTML]{008000}{\scriptsize{(+52.1\%)}} & \textbf{35.5} \textcolor[HTML]{008000}{\scriptsize{(+34.0\%)}} & \textbf{30.1} & \textbf{92.3} \textcolor[HTML]{008000}{\scriptsize{(+78.3\%)}} & \textbf{96.2} \textcolor[HTML]{008000}{\scriptsize{(+55.0\%)}} & \textbf{55.7} \textcolor[HTML]{008000}{\scriptsize{(+48.9\%)}} & \textbf{50.2} & \textbf{81.5} \textcolor[HTML]{008000}{\scriptsize{(+70.7\%)}} \\
    \bottomrule
    \end{tabular}
  }
\end{table*}

\subsection{Unified Anomaly Score}

Finally, we combine the visual and semantic signals into a single anomaly score:
\begin{equation}
S(x)
=
(1-\alpha)\,S_{\mathrm{vis}}(x)
+
\alpha\,S_{\mathrm{text}}(x),
\label{eq:final_score}
\end{equation}
where $\alpha \in [0,1]$ balances structural deviation and semantic inconsistency.

This final formulation reflects the central design principle of LAKE. An anomaly should not be identified solely as a geometric outlier, nor solely as a semantic label mismatch. Instead, reliable anomaly detection requires both: deviation from the normal feature manifold and activation of abnormal semantics. By integrating these two complementary cues, LAKE transforms anomaly detection from external task adaptation into latent knowledge excavation inside a frozen VLM. As summarized in Algorithm~\ref{alg:lake}, the entire pipeline requires only a small normal support set and inference-time feature probing, while preserving interpretability at both the neuron level and the patch level.

\section{Experiments and Analysis}

\subsection{Experimental Setup}

\noindent\textbf{Datasets.} To maintain fairness and consistency with prior arts, we adopt the multi-class joint evaluation protocol and data splitting strategy introduced by VisualAD~\cite{hou2026visualadlanguagefreezeroshotanomaly}. Our primary experiments are conducted on three widely recognized industrial anomaly detection benchmarks: MVTec-AD~\cite{bergmann2019mvtec}, VisA~\cite{zou2022spot}, and BTAD~\cite{btad2020}. Additionally, to rigorously assess the cross-domain generalizability of the LAKE framework, we extend our evaluation to the medical imaging domain using the Brain-AD~\cite{baid2021rsna} dataset.

\noindent\textbf{Evaluation Metrics.} Following standard practices in the field~\cite{hou2026visualadlanguagefreezeroshotanomaly}, we assess global image-level anomaly detection performance using the Area Under the Receiver Operating Characteristic curve (AUROC), Average Precision (AP), and the maximum F1-score ($F_1$-max) \cite{bergmann2019mvtec}. For fine-grained pixel-level anomaly localization, we evaluate the models using pixel-wise AUROC, AP, $F_1$-max, and the Per-Region Overlap (PRO) metric, which ensures that anomalous regions of varying sizes are weighted equally \cite{bergmann2020uninformed}.

\noindent\textbf{Hyperparameters.} As an inherently training-free framework, LAKE requires no gradient optimization or learnable parameters. We utilize the pre-trained CLIP model (ViT-L/14@336px) \cite{radford2021learning} as the visual backbone, with all input images uniformly resized to 336$\times$336 pixels. Textual embeddings are generated using the standardized prompt templates: "a photo of a normal [class]" and "a photo of an anomalous [class]". The normal reference gallery is constructed under a 64-shot setting. For anomaly-sensitive neuron detection, the dimension of the sensitive subspace is fixed at $K = 100$, and the balancing coefficient for cross-modal fusion is set to $\alpha = 0.3$. Finally, the coarse patch-level anomaly scores are upsampled to the original image resolution using bilinear interpolation to produce the fine-grained pixel-level anomaly maps. All experiments are executed on a single NVIDIA H200 GPU.

\subsection{Comparison with State-of-the-Art Methods}

In this study, we compare our LAKE framework with seven zero- and few-shot baselines: the native OpenCLIP backbone, WinCLIP~\cite{jeong2023winclip}, CLIP-AD~\cite{chen2024clip}, AnomalyCLIP~\cite{zhouanomalyclip}, AdaCLIP~\cite{cao2024adaclip}, and two variants of VisualAD~\cite{hou2026visualadlanguagefreezeroshotanomaly} (utilizing CLIP~\cite{radford2021learning} and DINOv2~\cite{oquab2023dinov2}). Furthermore, we incorporate the recent ReMP-AD~\cite{ma2025remp} as a competitive benchmark to further validate our performance gains."

\begin{figure*}[t]
\centering
\includegraphics[width=0.95\linewidth]{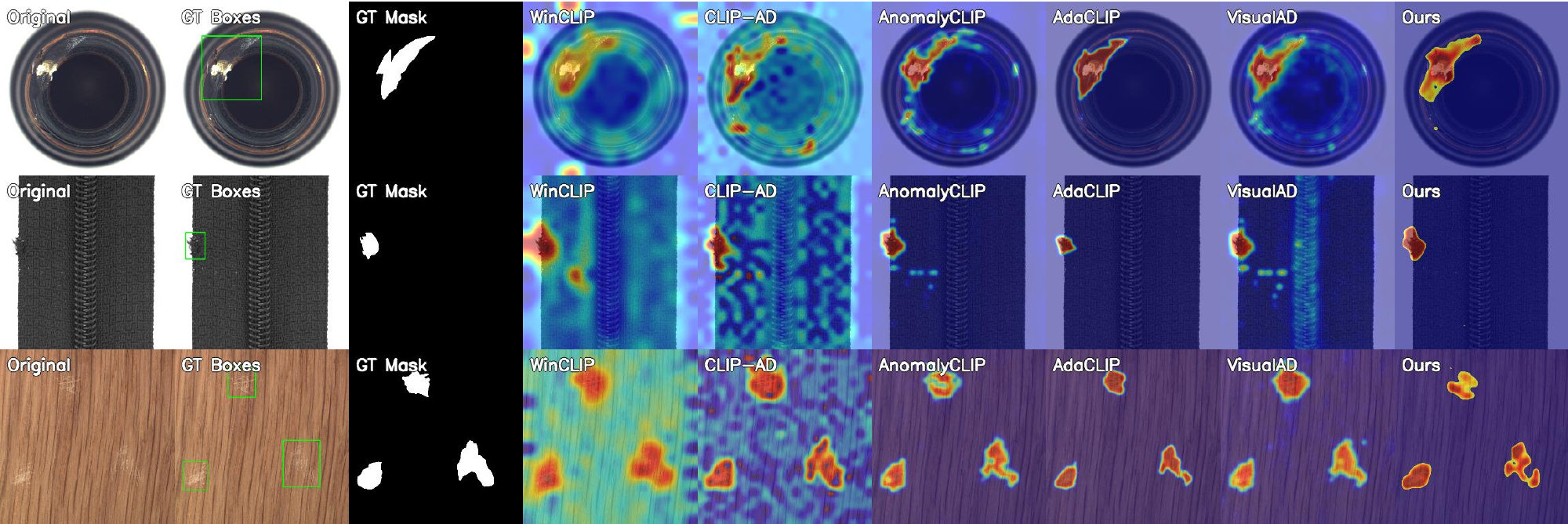} 
\caption{Qualitative comparison of anomaly localization with state-of-the-art (SOTA) methods.}
\label{fig:dingxing}
\end{figure*}

\begin{figure*}[t]
\centering
\includegraphics[width=0.95\linewidth]{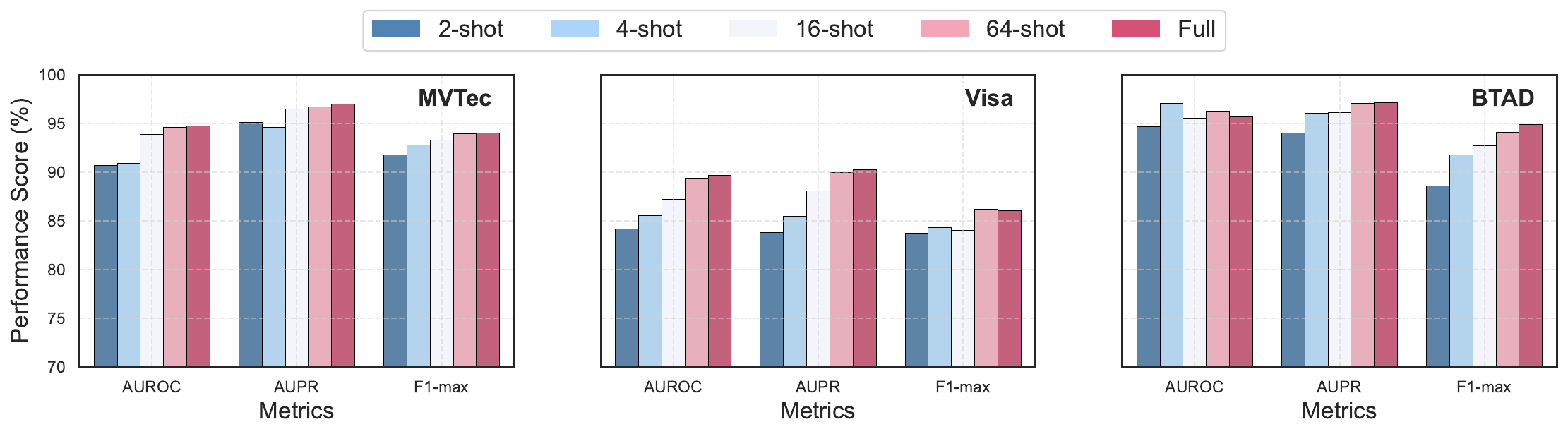} 
\caption{Ablation of support set size on image-level anomaly detection.}
\label{fig:image_set_ablation}
\end{figure*}

\begin{figure*}[t]
\centering
\includegraphics[width=0.95\linewidth]{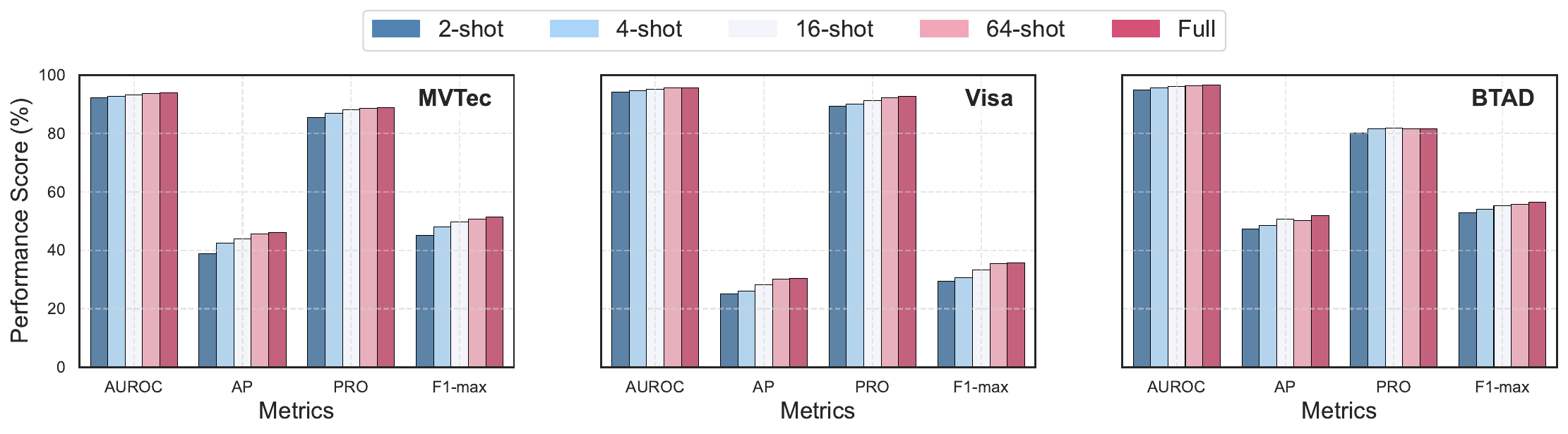} 
\caption{Ablation of support set size on pixel-level anomaly localization.}
\label{fig:pixel_set_ablation}
\end{figure*}

\noindent\textbf{Quantitative Results.} To comprehensively evaluate our proposed training-free framework, we conducted comparative experiments across the MVTec-AD, VisA, and BTAD benchmarks. The results demonstrate that LAKE establishes new SOTA performance in both global image-level anomaly detection (Table~\ref{tab:image_level_transposed}) and fine-grained pixel-level localization (Table~\ref{tab:pixel_level_transposed}). For image-level detection, our approach consistently outpaces the best-performing baselines, achieving significantly higher AUROC scores (e.g., 94.7\% on MVTec-AD versus VisualAD's 92.2\%, and 89.4\% on VisA versus VisualAD's 84.7\%). Furthermore, when compared to the naive OpenCLIP baseline, LAKE delivers massive performance leaps, such as a 70.5\% AUROC increase on the BTAD dataset, proving that explicit latent knowledge extraction is vastly superior to directly utilizing raw features.
This global superiority extends seamlessly to pixel-level anomaly localization. Our method consistently surpasses the strongest Per-Region Overlap (PRO) limits, achieving 88.9\% on MVTec-AD, 92.3\% on VisA, and 81.5\% on BTAD, outperforming models that rely on heavy downstream adaptation. Ultimately, these comprehensive gains validate our core hypothesis: high-precision anomaly detection does not strictly require external adapters or massive memory banks. Rather, by accurately isolating a sparse subset of anomaly-sensitive neurons, our framework intrinsically excavates and activates the latent anomaly knowledge already embedded within pre-trained foundation models, allowing it to seamlessly perceive both structural deviations and semantic mismatches.

\begin{figure*}[t]
\centering
\includegraphics[width=1\linewidth]{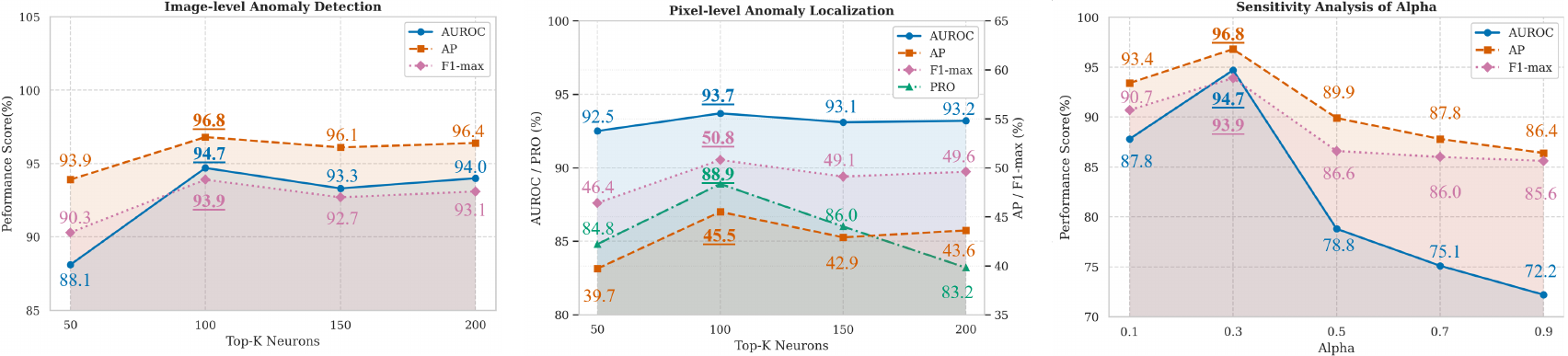} 
\caption{Ablation studies on key hyperparameters. Impact of the top-$k$ selection at the image level (Left). Effect of the top-$k$ selection at the pixel level (Middle). Performance variations across different values of the parameter $\alpha$ (Right).}
\label{fig:ablation}
\end{figure*}

\noindent\textbf{Qualitative Results.} To intuitively evaluate the fine-grained anomaly localization capability of our framework, we conducted a qualitative visual comparison against state-of-the-art baselines across diverse and complex industrial scenarios (Figure~\ref{fig:dingxing}). The visualizations demonstrate that LAKE consistently generates significantly sharper, cleaner, and more precise anomaly heatmaps. Specifically, early baselines such as WinCLIP and CLIP-AD suffer from highly diffuse activations and severe background false positives, struggling to isolate defects from complex underlying textures. While subsequent methods show improvement, our approach uniquely achieves tight alignment with the actual defect morphologies. It successfully suppresses task-irrelevant normal patterns and precisely delineates anomalous regions, even successfully localizing multiple scattered defects (as observed in the wood surface scenario) that often confound other models. Ultimately, this qualitative evidence confirms that by restricting feature analysis to a sparse subset of anomaly-sensitive neurons and integrating cross-modal textual probing to filter redundant dimensions, our framework intrinsically excavates robust latent anomaly knowledge rather than merely fitting to external annotations.

\subsection{Ablation Study}

\noindent\textbf{Support Set Size.} To investigate the data efficiency and robustness of our framework, we evaluated its performance across support set sizes ranging from extreme few-shot scenarios (2, 4, and 16 shots) to 64 shots, comparing it directly against the full-shot baseline. The results demonstrate that both global image-level anomaly detection (Figure~\ref{fig:image_set_ablation}) and fine-grained pixel-level localization (Figure~\ref{fig:pixel_set_ablation}) exhibit a rapid initial improvement as the number of support samples increases, but distinctively plateau at 64 shots. Specifically, key metrics in the 64-shot configuration, such as image-level AUROC, pixel-level AP, and PRO, are nearly indistinguishable from those achieved in the full-shot setting across the MVTec-AD, VisA, and BTAD benchmarks. This consistent stabilization underscores the extreme data efficiency of our training-free architecture. It proves that a modest number of normal samples is entirely sufficient to accurately approximate the normal feature manifold and activate the critical anomaly-sensitive neurons, fundamentally bypassing the need for traditional data-hungry downstream learning paradigms.

\noindent\textbf{Top-K Anomaly-Sensitive Neurons.} We ablate the subspace dimension $K$ on MVTec-AD to validate our sparsity hypothesis. As shown in Figure~\ref{fig:ablation} (Left, Middle), LAKE achieves optimal image- and pixel-level performance at $K=100$ (e.g., 94.7\% image AUROC, 88.9\% PRO). Overconstraining the subspace ($K=50$) discards critical normal variations, while excessive expansion ($K \ge 150$) introduces redundant, task-irrelevant patterns that dilute anomalous signals. This confirms that latent anomaly knowledge is highly concentrated within a sparse, sensitive neuronal subset rather than uniformly distributed. Consequently, mathematically anchoring this optimal sparsity allows our framework to effectively bypass the massive feature redundancy typical of foundation models.

\noindent\textbf{Text-Weighting Parameter.} We evaluate the cross-modal fusion weight $\alpha$ for image-level detection on MVTec-AD (Figure~\ref{fig:ablation} Right), keeping pixel-level localization strictly visual to preserve fine-grained boundaries. Results indicate that visual deviations must dominate for optimal performance, peaking at $\alpha=0.3$. Assigning excessive weight to textual semantics ($\alpha \ge 0.5$) precipitously degrades accuracy, while insufficient weighting ($\alpha=0.1$) fails to fully exploit cross-modal reasoning. This demonstrates that while textual prompts provide valuable auxiliary verification, the core driver of zero-shot anomaly detection remains the structural knowledge extracted from sparse sensitive neurons. Ultimately, this carefully calibrated fusion ensures that semantic alignment acts as a robust filter, preventing generalized language priors from overriding the model's intrinsic geometric perception.

\subsection{Mechanistic Interpretability and Analysis}

\begin{table}[t]
  \centering
  \caption{Quantitative evaluation of neuron selection: Top-K vs. Random selection strategies on MVTec-AD under 64-shot.}
  \label{tab:neuron_selection}
  \renewcommand{\arraystretch}{1.1}
  \resizebox{\linewidth}{!}{
    \begin{tabular}{l ccc c}
    \toprule
    \textbf{Model} & \textbf{AUROC} $\uparrow$ & \textbf{$F_1$-max} $\uparrow$ & \textbf{AP} $\uparrow$ & \textbf{PRO} $\uparrow$ \\
    \midrule
    \multicolumn{5}{c}{\textit{Image-Level Performance}} \\
    \midrule
    LAKE (Random 100) & 81.6
 & 83.2
 & 83.5
 & - \\
       \rowcolor[HTML]{E5F0F7}
    LAKE (Top 100)  & \textbf{94.7
} & \textbf{93.9
} & \textbf{96.7
} & - \\
    
    \midrule
    \multicolumn{5}{c}{\textit{Pixel-Level Performance}} \\
    \midrule
    LAKE (Random 100) & 78.5
 & 20.4
 & 20.0
 & 45.8
 \\
        \rowcolor[HTML]{E5F0F7}
    LAKE (Top 100)  & \textbf{93.7
} & \textbf{50.8
} & \textbf{45.5
} & \textbf{88.9
} \\
    \bottomrule
    \end{tabular}}
\end{table}

\begin{figure}[t]
\centering
\includegraphics[width=0.98\linewidth]{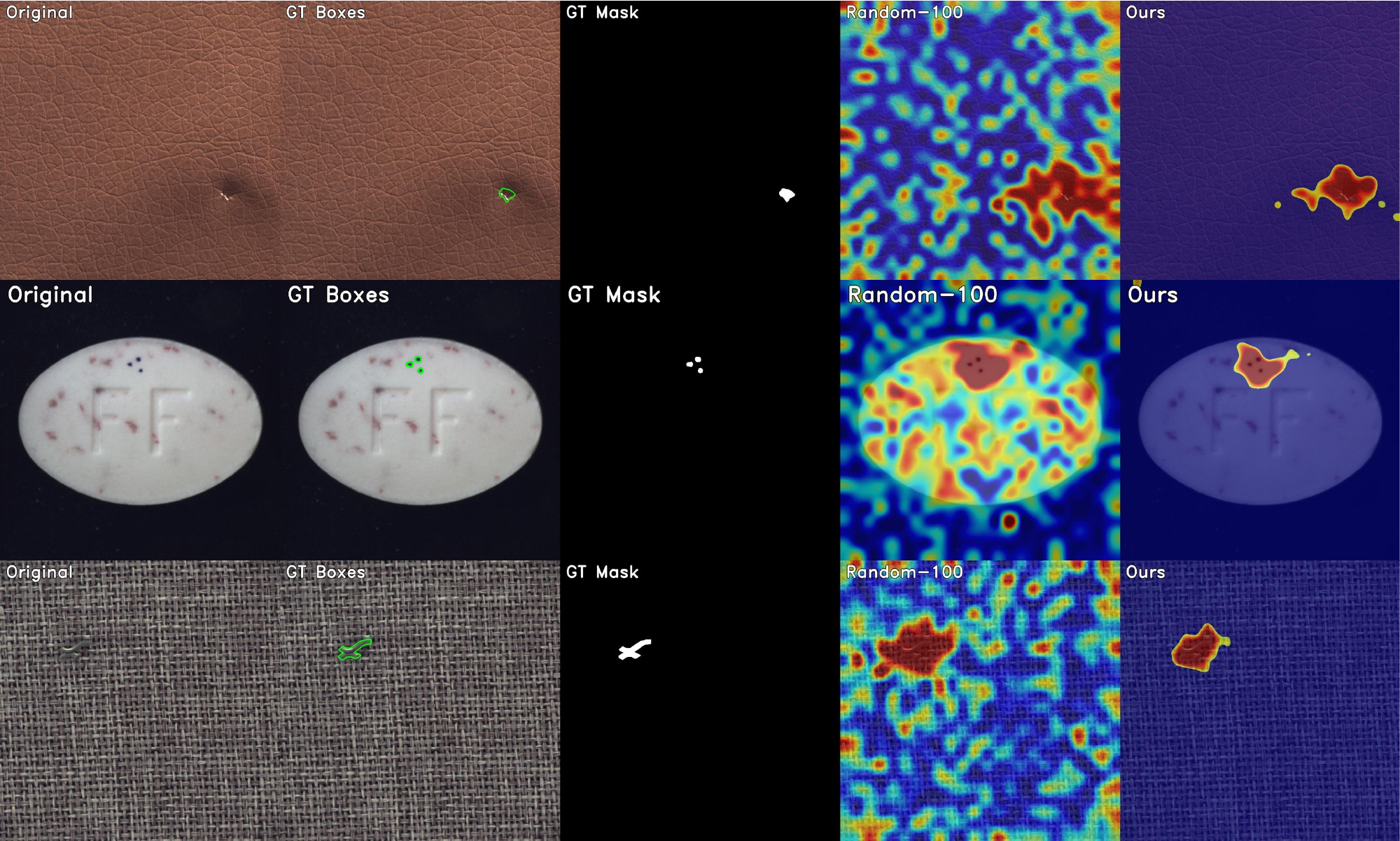}
\caption{{Qualitative evaluation of neuron specificity for anomaly excavation.} Heatmaps generated by random neurons exhibit severe background noise, whereas our explicitly identified anomaly-sensitive neurons yield highly precise localizations that tightly align with the ground truth.}
\label{fig:dingxing_neuron}
\end{figure}

\noindent\textbf{Quantitative Analysis of Neuron Specificity.} To verify the task-specific importance of the identified neurons, we compare our excavated Top-100 neurons against 100 randomly selected neurons from the same layer (Table~\ref{tab:neuron_selection}). The results confirm that anomaly-discriminative capabilities are highly concentrated within a sparse subset rather than uniformly distributed. While random selection introduces task-irrelevant background noise that severely degrades performance, our Top-100 neurons achieve a massive leap in precision. Notably, they improve the image-level AUROC from 81.6\% to 94.7\% and nearly double the fine-grained pixel-level PRO score from 45.8\% to 88.9\%. This provides compelling evidence that the LAKE framework accurately localizes and activates intrinsic latent anomaly knowledge rather than relying on spurious correlations.

\noindent\textbf{Qualitative Analysis of Neuron Specificity.} To intuitively evaluate the impact of neuron selection, we visually compare the heatmaps generated by our Top-100 neurons against the random baseline. As shown in Figure~\ref{fig:dingxing_neuron}, random selection produces diffuse, unfocused activations that completely fail to separate anomalies from complex backgrounds. In stark contrast, the heatmaps driven by our anomaly-sensitive neurons tightly align with actual defect morphologies and successfully suppress task-irrelevant normal patterns. These visualizations directly corroborate our quantitative findings, confirming that the precise localization of sparse sensitive neurons is vital for robust, interpretable anomaly excavation.

\begin{figure}[t]
\centering
\includegraphics[width=0.9\linewidth]{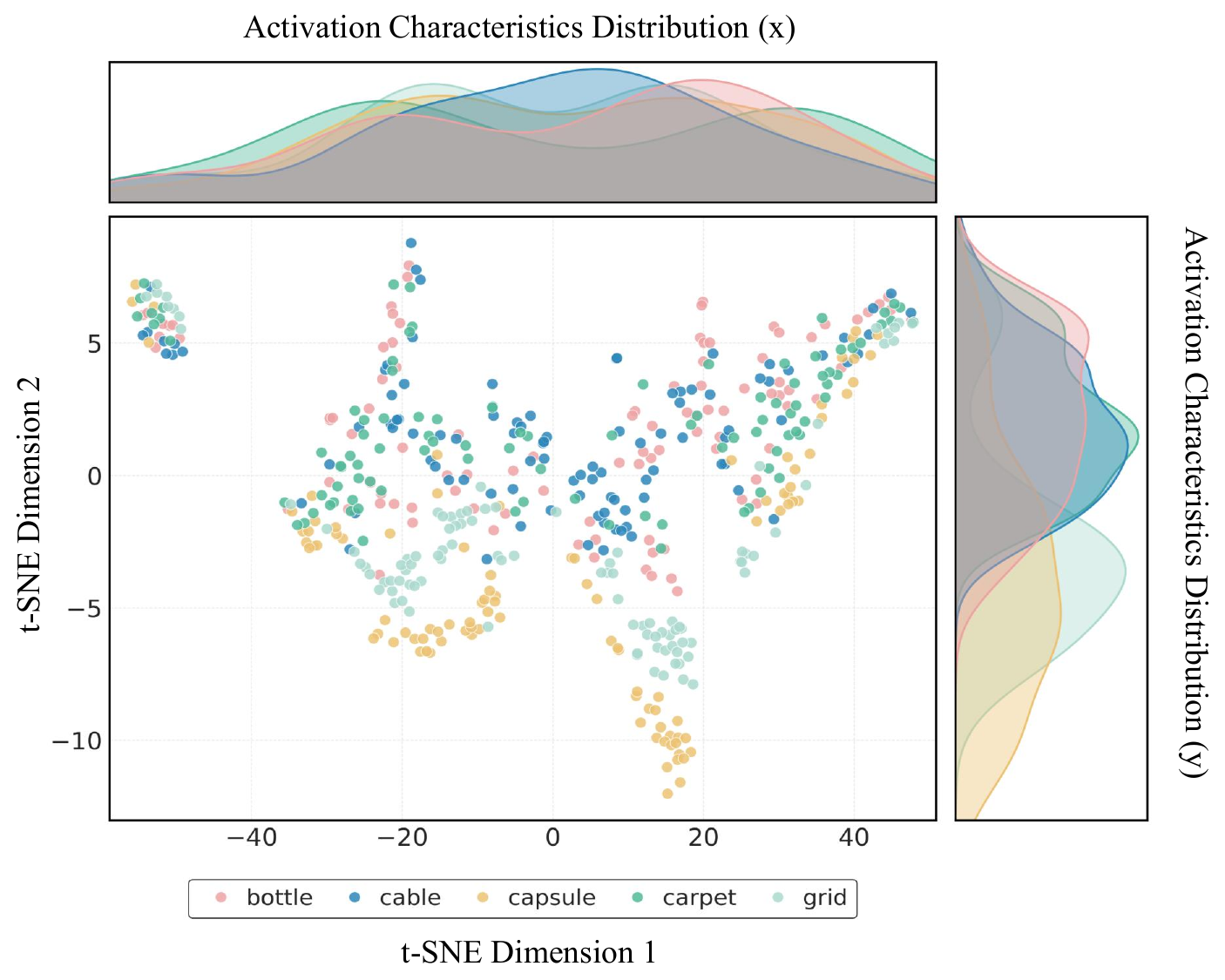}
\caption{{Cross-category overlap analysis.} The t-SNE demonstrates the significant spatial overlap of identified anomaly-sensitive neurons across five diverse MVTec-AD categories, highlighting their generalized activation patterns.}
\label{fig:chongdie}
\end{figure}

\noindent\textbf{Cross-category Neuron Overlap Analysis.}
To verify the cross-scenario generality of the mined neurons, we conducted a cross-category neuron overlap analysis. As shown in Figure~\ref{fig:chongdie}, results show that the key neurons exhibit strong consistency and generality in spatial distribution across distinct industrial categories. Specifically, t-SNE visualization of neuronal fingerprints from five structurally different categories (e.g., Bottle, Cable) reveals that scatter points are highly intertwined rather than forming isolated clusters. Furthermore, the nearly overlapping marginal probability density curves quantitatively demonstrate the high similarity in activation characteristics. Therefore, this proves that the mined neurons represent a universal cross-category anomaly semantic concept, providing microscopic evidence for the model's robust zero-shot generalization.

\begin{table}[t]
  \centering
  
  \caption{Performance Comparison at Image and Pixel Levels. The best results are highlighted in bold.}
  \label{tab:performance}
  \renewcommand{\arraystretch}{1.1}
  \resizebox{\linewidth}{!}{
    \begin{tabular}{l ccc c}
    \toprule
    \textbf{Model} & \textbf{AUROC} $\uparrow$ & \textbf{$F_1$-max} $\uparrow$ & \textbf{AP} $\uparrow$ & \textbf{PRO} $\uparrow$ \\
    \midrule
    \multicolumn{5}{c}{\textit{Image-Level Performance}} \\
    \midrule
    WinCLIP~\cite{jeong2023winclip}         & 90.4 & 92.7 & 95.6 & - \\
    \rowcolor[HTML]{E5F0F7}
    + Ours  & \textbf{92.8} {\scriptsize \textcolor[HTML]{008000}{(+2.4\%)}} & \textbf{93.4} {\scriptsize \textcolor[HTML]{008000}{(+0.7\%)}} & \textbf{95.8} {\scriptsize \textcolor[HTML]{008000}{(+0.2\%)}} & - \\
    \midrule
    ReMP-AD~\cite{ma2025remp}         & 96.8 & 95.5 & --   & - \\
    \rowcolor[HTML]{E5F0F7}
    + Ours  & \textbf{96.9} {\scriptsize \textcolor[HTML]{008000}{(+0.1\%)}} & \textbf{96.2} {\scriptsize \textcolor[HTML]{008000}{(+0.7\%)}} & \textbf{96.9} & - \\
    
    \midrule
    \multicolumn{5}{c}{\textit{Pixel-Level Performance}} \\
    \midrule
    WinCLIP~\cite{jeong2023winclip}          & 82.3 & 24.8 & 18.2 & 62.0 \\
    \rowcolor[HTML]{E5F0F7}
    + Ours  & \textbf{95.7} {\scriptsize \textcolor[HTML]{008000}{(+13.4\%)}} & \textbf{51.9} {\scriptsize \textcolor[HTML]{008000}{(+27.1\%)}} & \textbf{48.3} {\scriptsize \textcolor[HTML]{008000}{(+30.1\%)}} & \textbf{89.0} {\scriptsize \textcolor[HTML]{008000}{(+27.0\%)}} \\
    \midrule
    ReMP-AD~\cite{ma2025remp}          & 96.6 & 61.1 & --   & 92.8 \\
    \rowcolor[HTML]{E5F0F7}
    + Ours  & \textbf{96.9} {\scriptsize \textcolor[HTML]{008000}{(+0.3\%)}} & \textbf{63.5} {\scriptsize \textcolor[HTML]{008000}{(+2.4\%)}} & \textbf{61.8} & \textbf{93.9} {\scriptsize \textcolor[HTML]{008000}{(+1.1\%)}} \\
    \bottomrule
    \end{tabular}}
\end{table}

\subsection{Extensibility and Generalization}
\noindent\textbf{Plug-and-Play Integration.}
To verify the plug-and-play characteristics and universal enhancement capabilities of our proposed method, we integrated it as an independent module into mainstream anomaly detection models (WinCLIP~\cite{jeong2023winclip} and ReMP-AD~\cite{ma2025remp}) and evaluated their performance. The results demonstrate that our method effectively adapts to different baselines and delivers consistent, significant performance improvements across both image and pixel levels (Table~\ref{tab:performance}). Specifically, integrating our method into WinCLIP yields a breakthrough in pixel-level AUROC (from 82.3\% to 95.7\%) alongside image-level AUROC gains (from 90.4\% to 92.8\%), while on the already high-performing ReMP-AD baseline, it still provides effective gains, such as increasing image-level $F_1$-max from 95.5\% to 96.2\% and pixel-level AUROC from 92.8\% to 96.9\%. Therefore, this proves that our latent anomaly knowledge excavation mechanism serves as an efficient, universal plug-and-play module that significantly activates the intrinsic potential of existing vision-language models for zero-shot anomaly perception without requiring complex downstream fine-tuning.

\begin{table}[t]
  \centering
  \caption{Anomaly detection performance on Brain-AD. The best results are highlighted in bold.}
  \label{tab:brain}
  \renewcommand{\arraystretch}{1.1}
  \resizebox{\linewidth}{!}{
    \begin{tabular}{l ccc c}
    \toprule
    \textbf{Method} & \textbf{AUROC} $\uparrow$ & \textbf{$F_1$-max} $\uparrow$ & \textbf{AP} $\uparrow$ & \textbf{PRO} $\uparrow$ \\
    \midrule
    \multicolumn{5}{c}{\textit{Image-Level Performance}} \\
    \midrule
    WinCLIP~\cite{jeong2023winclip}          & 72.7 & 90.7 & 91.6 & - \\
    CLIP-AD~\cite{chen2024clip}              & 72.1 & 91.3 & 91.5 & - \\
    AnomalyCLIP~\cite{zhouanomalyclip}       & 69.0 & 90.6 & 90.1 & - \\
    AdaCLIP~\cite{cao2024adaclip}            & 80.0 & 90.2 & 94.1 & - \\
    VisualAD (CLIP)~\cite{hou2026visualadlanguagefreezeroshotanomaly}  & 80.8 & 91.8 & 94.7 & - \\
    VisualAD (DINOv2)~\cite{hou2026visualadlanguagefreezeroshotanomaly}& 87.1 & 92.5 & 96.7 & - \\
    \midrule
    \rowcolor[HTML]{E5F0F7}\textbf{LAKE (Ours)}    & \textbf{87.4} & \textbf{95.7} & \textbf{98.6} & - \\
    
    \midrule
    \multicolumn{5}{c}{\textit{Pixel-Level Performance}} \\
    \midrule
    WinCLIP~\cite{jeong2023winclip}          & 87.6 & 21.7 & 13.3 & 59.7 \\
    CLIP-AD~\cite{chen2024clip}              & 94.1 & 42.9 & 40.5 & 74.2 \\
    AnomalyCLIP~\cite{zhouanomalyclip}       & 95.1 & 43.1 & 42.3 & 71.5 \\
    AdaCLIP~\cite{cao2024adaclip}            & 95.2 & 40.6 & 37.0 & 36.4 \\
    VisualAD (CLIP)~\cite{hou2026visualadlanguagefreezeroshotanomaly}  & 95.2 & 46.7 & 43.7 & 79.5 \\
    VisualAD (DINOv2)~\cite{hou2026visualadlanguagefreezeroshotanomaly}& 96.4 & 50.2 & 51.9 & 83.5 \\
    \midrule
    \rowcolor[HTML]{E5F0F7}\textbf{LAKE (Ours)}    & \textbf{97.2} & \textbf{50.9} & \textbf{52.0} & \textbf{85.3} \\
    \bottomrule
    \end{tabular}}
\end{table}

\noindent\textbf{Cross-Domain Generalization.}
To evaluate generalizability beyond standard industrial scenarios, we conducted experiments on the medical anomaly dataset, Brain-AD~\cite{baid2021rsna}. The results demonstrate that our method exhibits exceptional cross-domain transferability, achieving state-of-the-art performance in key metrics for both global detection and fine-grained localization (Table~\ref{tab:brain}). Specifically, at the image level, our framework achieves an unparalleled AP of 98.6\% and an F1-max of 95.7\% (significantly outperforming the strong VisualAD (DINOv2) baseline), while at the pixel level, it yields the highest AP of 52.0\% and a leading PRO score of 85.3\%, alongside highly competitive pixel-level AUROC (97.2\%) and F1-max (50.9\%) scores. Therefore, this strongly validates that the latent anomaly knowledge excavated by our anomaly-sensitive neurons is not overfit to industrial textures, but rather captures a universal, domain-agnostic understanding of abnormality that generalizes effectively to distinct fields such as medical imaging.

\section{Conclusion}

In this paper, we introduce LAKE, a training-free framework that elicits latent anomaly knowledge from sparse anomaly-sensitive neurons in pre-trained VLMs. By localizing these neurons through geometric variance and cross-modal probing, LAKE seamlessly integrates visual structural deviations with textual mismatch signals. Extensive experiments confirm that our method achieves state-of-the-art performance across multiple industrial benchmarks and exhibits exceptional cross-domain generalizability to fields such as medical imaging. Ultimately, this work offers a transparent and highly efficient paradigm by redefining anomaly detection as the targeted activation of intrinsic model knowledge.

\bibliographystyle{IEEEtran}
\bibliography{main.bib}

\clearpage

\begin{appendices}

\section*{Appendix}
The appendices of this paper provide a comprehensive technical extension of the LAKE framework, beginning with Appendix~\ref{text}, which details a minimalist and consistent text prompting strategy that employs fixed templates for normal and abnormal states to ensure training-free integrity. Appendix~\ref{compute} demonstrates the system's efficiency, highlighting how projecting features into a low-dimensional subspace achieves over 90\% memory compression and enables real-time inference at approximately 38.3 FPS. In Appendix~\ref{theo}, the authors provide mathematical justifications, proving that their variance-based neuron selection is equivalent to Truncated PCA and that max-pooling is theoretically superior for detecting sparse, localized defects. Appendix~\ref{dis} addresses practical implementation concerns through a Q\&A format, explaining the interpretability of high-variance neurons and the framework's robustness against sample contamination. Finally, Appendix~\ref{lim} acknowledges current limitations, noting that while LAKE excels in 2D image anomaly detection, its application to temporal video sequences or 3D point clouds remains an objective for future research.

\section{Text Prompt Construction}
\label{text}

To maintain the 'training-free' integrity of our framework, we employ a minimalist prompting strategy. We avoid the common practice of prompt ensembling (averaging multiple templates), which can artificially inflate performance through heavy manual engineering. For any given category $c$, we define the state-specific text embeddings $t_{norm}$ and $t_{anom}$ using the following templates:\begin{itemize}\item \texttt{Normal: "a photo of a normal [class]."}\item \texttt{Abnormal: "a photo of an anomalous [class]."}\end{itemize} These templates remain constant across all datasets (MVTec-AD~\cite{bergmann2019mvtec}, VisA~\cite{zou2022spot}, Brain-AD~\cite{baid2021rsna}). This consistency demonstrates that LAKE does not require category-specific prompt tuning to achieve state-of-the-art results.

\section{Computational Complexity and Efficiency}
\label{compute}

\noindent\textbf{Memory Efficiency.}
To address the memory bottleneck in methods like PatchCore, where space complexity $\mathcal{O}(N_{support} \times N_{patches} \times D)$ scales prohibitively with high-dimensional features ($D \ge 896$), we propose a compact Normality Representation powered by an Anomaly-Sensitive Neurons selection mechanism. 

By projecting features into a discriminative low-dimensional subspace ($K=100$), we achieve a compression rate of nearly $90\%$ while filtering out anomaly-insensitive noise. Empirical evaluations demonstrate that our approach stabilizes global peak VRAM at approximately 1.5 GB, with online inference requiring only 1229.28 MB, offering a significant memory advantage over traditional memory-bank methods that typically consume several gigabytes.

\noindent\textbf{Time Efficiency.}
Beyond space complexity, our training-free method achieves exceptional Inference Latency by decoupling gallery construction into one-time offline operations, thereby imposing zero burden on runtime throughput. During the online phase, building the Image Memory Bank under a 64-shot setup takes a mere 0.55 ms, while end-to-end Single Image Inference is clocked at 25.11 ms (approx. 39.8 FPS). This efficiency is primarily driven by our Visual Scouring stage, where projecting features into a low-dimensional subspace ($K=100$) reduces the complexity of Nearest-Neighbor Search from $\mathcal{O}(D)$ to $\mathcal{O}(K)$, effectively circumventing the curse of dimensionality. By combining this dimensionality reduction with optimized cross-modal alignment, our approach significantly outpaces high-dimensional retrieval methods like PatchCore and matches the latency benchmarks of lightweight models like WinCLIP~\cite{jeong2023winclip}, offering a highly competitive paradigm for efficient multi-modal anomaly detection.

\section{Theoretical Justifications}
\label{theo}
In this section, we provide rigorous theoretical insights into the core mathematical designs of the LAKE framework, demonstrating that our seemingly heuristic operations are firmly grounded in manifold learning, probabilistic inference, and extreme value theory.

\subsection{Variance Selection as Truncated PCA}

In Section 3.2, we hypothesize that channels with high variance under normal data approximate the principal axes of the expected normality. Here, we formally justify this by bridging our approach with principal component analysis (PCA) on feature manifolds.

Let the feature representation of normal patches at layer $l$ be a random vector $h \in \mathbb{R}^D$. We assume that normal features reside on a low-dimensional manifold $\mathcal{M}$ embedded in the high-dimensional space. The intrinsic structure of this normal manifold can be captured by its covariance matrix:
\begin{equation}
    \Sigma = \mathbb{E}[(h - \mu)(h - \mu)^\top],
\end{equation}
where $\mu = \mathbb{E}[h]$ is the mean vector. To project the data into a subspace that maximizes the retained structural information (variance), traditional PCA seeks the eigenvectors of $\Sigma$ corresponding to the largest eigenvalues.

In modern deep Vision-Language Models (VLMs) like CLIP, the high-dimensional activation space is highly disentangled, meaning that the feature dimensions are largely uncorrelated. Under this widely accepted assumption, the covariance matrix $\Sigma$ is strictly diagonally dominant:
\begin{equation}
    \Sigma \approx \text{diag}(\sigma_1^2, \sigma_2^2, \dots, \sigma_D^2).
\end{equation}
Because $\Sigma$ is diagonal, its eigenvectors are simply the standard basis vectors $e_d \in \mathbb{R}^D$, and the corresponding eigenvalues are precisely the channel variances $\lambda_d = \sigma_d^2$.

Therefore, explicitly decomposing the covariance matrix is mathematically redundant. Ranking the channels by their marginal variances $\sigma_d^2$ (as done in Eq. 3) and selecting the Top-$K$ dimensions (Eq. 4) is mathematically equivalent to performing Truncated PCA on the local tangent space of the normal manifold $\mathcal{M}$. This proves that our anomaly-sensitive subspace $I_{sens}$ rigorously captures the principal orthogonal components of normal visual patterns without the computational overhead of singular value decomposition (SVD).

\subsection{Extremal Value Guarantee of Max-Pooling for Sparse Defects}

In Eq. 8, we apply a token-wise max-pooling operation $S_{vis}(x) = \max_{i} d_i(x)$ to aggregate patch-level deviations. This design is critical for industrial anomaly detection, where defects are typically highly localized and sparse. We can justify this mathematically using Extreme Value Theory.

Suppose an image contains $N$ patch tokens. Let $m$ be the number of anomalous tokens, where $m \ll N$. Let the distance scores $d_i$ for normal patches follow a distribution with mean $\mu_0$, and anomalous patches follow a distribution with mean $\mu_1$, where $\mu_1 \gg \mu_0$.

If we were to use global average pooling (GAP), the expected image-level score would be:
\begin{equation}
    \mathbb{E}[S_{avg}] = \frac{1}{N} \left( \sum_{i \in \text{normal}} d_i + \sum_{j \in \text{anom}} d_j \right) \approx \frac{N-m}{N}\mu_0 + \frac{m}{N}\mu_1.
\end{equation}
As image resolution grows ($N \to \infty$) and the anomaly remains small ($m$ is constant), $\lim_{N \to \infty} \mathbb{E}[S_{avg}] = \mu_0$. The anomalous signal $\mu_1$ is completely diluted by the dominant normal background, causing false negatives.

Conversely, with Max-Pooling, the final score $S_{max}$ is determined by the highest deviation. By definition, the maximum over all patches is strictly lower-bounded by the maximum over the anomalous patches:
\begin{equation}
    S_{max} = \max_{1 \le i \le N} d_i \ge \max_{j \in \text{anom}} d_j \approx \mu_1.
\end{equation}
Therefore, $S_{max}$ preserves the extremal anomalous signal independently of the background size $N$, guaranteeing theoretical robustness for detecting minuscule defects.

\section{More Discussions}
\label{dis}
\noindent$\triangleright$ \textbf{\textit{Q1. Why use high-variance neurons when anomalies traditionally lie in low-variance residual spaces?}}

Unlike traditional methods, the highly disentangled nature of VLMs renders low-variance channels as dormant, task-irrelevant noise. Conversely, high-variance channels continuously encode the core components of expected normality, mathematically acting as the principal axes of the normal manifold (Appendix C.1). Since anomalies inherently disrupt this established normal manifold, their presence manifests as salient geometric deviations along these active, high-variance directions rather than in dormant channels.

\noindent$\triangleright$ \textbf{\textit{Q2. Does requiring 64 normal samples violate the claim of a ``training-free'' framework?}}

No. In the context of foundation models, ``training'' specifically refers to gradient-based backpropagation, parameter updates, or iterative optimization (such as adapter or soft-prompt tuning). Our framework requires none of these, keeping the underlying VLM strictly frozen. The 64 normal samples are utilized solely for a one-time, deterministic statistical extraction, computing channel variance and constructing a reference gallery. This acts as an offline memory cache rather than a learned optimization process. This approach completely aligns with established few-shot anomaly detection protocols (akin to memory-bank construction in methods like PatchCore), ensuring that the entire inference pipeline remains rigorously training-free.

\noindent$\triangleright$ \textbf{\textit{Q3. How can we determine hyperparameters $K$ and $\alpha$ in real-world scenarios without an anomalous validation set?}}

LAKE avoids dataset-specific fitting by relying on universal VLM properties. Empirically, our defaults ($K=100$, $\alpha=0.3$) achieved SOTA across four highly diverse datasets (industrial and medical), proving their domain-agnostic robustness. Theoretically, both parameters can be estimated using {only} normal data: $K$ can be determined via the ``elbow method'' on the normal support set's variance decay curve (equivalent to Truncated PCA, Appendix C.1). Similarly, since physical defects primarily disrupt geometric structures rather than semantics, keeping $\alpha < 0.5$ naturally establishes textual semantics as a secondary verification filter.

\noindent$\triangleright$ \textbf{\textit{Q4. Will performance collapse if a few anomalous images contaminate the 64 normal support samples?}}

No, LAKE is inherently robust to minor contamination due to the spatial sparsity of anomalies. A few contaminated images yield a negligible fraction of anomalous tokens, ensuring the maximum variance statistics (used to identify the top-$K$ axes) remain dominated by the normal background. Furthermore, the vast majority of pure normal tokens in the reference gallery ensures robust nearest-neighbor matching, preventing false positives. For extreme contamination scenarios, our compact subspace ($K=100$) enables the computationally trivial integration of off-the-shelf outlier filters (e.g., Isolation Forest) offline to guarantee manifold purity.

\noindent$\triangleright$ \textbf{\textit{Q5. What is the fundamental advantage of your variance selection over standard PCA or SVD?}}

Beyond bypassing the prohibitive $O(D^3)$ computational bottleneck of standard SVD, our method uniquely preserves {mechanistic interpretability}. While PCA projects data into opaque linear combinations that destroy individual feature meanings, LAKE explicitly retains original active neurons to maintain an unentangled representation. This is structurally critical for our cross-modal activation: because visual and textual embeddings align strictly in the original latent space, mixing visual channels via SVD would shatter this cross-modal correspondence, rendering direct semantic verification impossible.

\section{Limitations}
\label{lim}

While LAKE achieves state-of-the-art performance in 2D image anomaly detection, its current scope is inherently bounded by the static and spatial nature of foundational vision-language models (VLMs) like CLIP~\cite{radford2021learning}. Because our framework relies on excavating knowledge from these pre-trained spaces, it does not natively account for temporal dynamics or 3D geometric variations. Extending this training-free mechanism to video anomaly detection or 3D point clouds necessitates adapting the method to Video- or 3D-Language Models. However, we believe our core philosophy, activating latent anomaly knowledge via sensitive neurons, is fundamentally transferable. Future work will explore excavating temporal- and depth-sensitive neurons within these specialized architectures to broaden the framework's multi-modal applicability.

\end{appendices}
\end{document}